\documentclass{article}

\usepackage{PRIMEarxiv}
\usepackage{microtype}
\usepackage{graphicx}
\usepackage{subfigure}
\usepackage{booktabs} 
\usepackage{xcolor}
\usepackage{physics}
\usepackage{amsfonts}
\usepackage{amssymb}
\usepackage{bbm}
\usepackage{cite}
\usepackage[ruled]{algorithm2e}
\usepackage{bm}
\usepackage{amsmath,amssymb,amsfonts}
\usepackage{algorithmic}

\usepackage[T1]{fontenc}

\usepackage[ruled]{algorithm2e}
\usepackage{bm}

\usepackage{longtable}
\usepackage{cite}
\usepackage{amsmath,amssymb,amsfonts}
\usepackage{algorithmic}
\usepackage{graphicx}
\usepackage{textcomp}
\usepackage{xcolor}
\usepackage{bm}
\def\BibTeX{{\rm B\kern-.05em{\sc i\kern-.025em b}\kern-.08em
    T\kern-.1667em\lower.7ex\hbox{E}\kern-.125emX}}

\usepackage{amsthm}
\usepackage{multirow}

\usepackage{caption}
\usepackage{url}            
\usepackage{booktabs}       
\usepackage{amsfonts}       
\usepackage{nicefrac}       
\usepackage{microtype}      
\usepackage{lipsum}
\usepackage{fancyhdr}       
\usepackage{graphicx}       
\graphicspath{{media/}}     

\usepackage{orcidlink}

\usepackage{hyperref}

\hypersetup{
  colorlinks=false,
  linkbordercolor=white,
 urlbordercolor=white,
pdfborder={0 0 0}
}

\title{AD-DMKDE: Anomaly Detection through Density Matrices and Fourier Features
\thanks{\textit{\underline{Citation}}: 
\textbf{Joseph et al., AD-DMKDE: Anomaly Detection through Density Matrices and Fourier Features.}} 
}

\author{
   Oscar A. Bustos-Brinez \orcidlink{0000-0003-0704-9117}, Joseph A. Gallego-Mejia \orcidlink{0000-0001-8971-4998}, Fabio A. Gonz\'{a}lez\orcidlink{0000-0001-9009-7288} \\
  MindLab \\
  Universidad Nacional de Colombia \\
  Bogot\'{a}, Colombia\\
  \texttt{\{oabustosb,jagallegom,fagonzalezo\}@unal.edu.co} 
}

\begin{document}
\maketitle

\begin{abstract}
This paper presents a novel density estimation method for anomaly detection using density matrices (a powerful mathematical formalism from quantum mechanics) and Fourier features. The method can be seen as an efficient approximation of Kernel Density Estimation (KDE). A systematic comparison of the proposed method with eleven state-of-the-art anomaly detection methods on various data sets is presented, showing competitive performance on different benchmark data sets. The method is trained efficiently and it uses optimization to find the parameters of data embedding. The prediction phase complexity of the proposed algorithm is constant relative to the training data size, and it performs well in data sets with different anomaly rates. Its architecture allows vectorization and can be implemented on GPU/TPU hardware. 
\end{abstract}

\vspace{8pt}

\keywords{density matrix \and random features \and anomaly detection \and quantum machine learning.}

\maketitle

\section{Introduction}
\label{sec:introduction}

An anomaly can be broadly defined as an observation or datum that deviates significantly from the patterns of the data set from which it originates, in one or more features. In most cases, data are generated by complex processes and allow for different types of measurements, so an anomaly may contain valuable information about anomalous behaviors of the generative processes, or elements that are impacting the generation or measurement stages \cite{aggarwal2016outlier}. Then, recognizing this type of data (which can be referred to as unusual, atypical, unexpected, malicious or rare, depending on the scenario), discerning real, meaningful anomalies from normal noisy samples (known as `outliers') and identifying the unusual processes that originate them are the main objectives of Anomaly Detection (AD) \cite{blazquez2021review}. Methods and algorithms that perform AD are key in various applications such as bank fraud detection, identification of abrupt changes in sensors, medical diagnostics, natural sciences, network security intrusion detection, among many others \cite{Ruff2021}.

However, classic AD methods face significant challenges that limit their performance and range of application \cite{pang2021deep}. In most cases, anomalies are much more sparse and scarce than normal data, and are not identified a priori as anomalies, which makes the use of supervised classifiers difficult. Furthermore, the boundaries between ``normal" and ``anomaly" regions are very subjective and depend on the particularities of each case. Data may also be contaminated by noise, so that both types of data may be intermingled, blurring the very idea of separation. Many of the AD algorithms address some of these difficulties, but at the cost of being vulnerable to others; however, the combination of classical methods with a good mathematical foundation and deep learning algorithms is one of the most promising paths \cite{Liu2020}, although it also presents shortcomings on the detection of complex types of anomalies and fewer possibilities of explanation.

The core idea of the proposed method is to use random Fourier features to approximate a Gaussian kernel centered in each training sample and then use a density matrix to summarize these kernels in an efficient and compact way. The density matrix is then used to estimate the density of new samples, and those whose density lies below a certain threshold are classified as anomalies. The method uses optimization to obtain the best random Fourier feature embedding (a process called ``Adaptive Fourier Features"), and is able to calculate the best threshold using percentiles from a validation data set.

The outline of the paper is as follows: in Section 2, we describe anomaly detection and density estimation in more depth, and present the benchmark methods to which we will compare our algorithm. In Section 3, we present the novel method, explaining the stages of the algorithm and how it uses Fourier features and density matrices. In Section 4, we systematically compare the proposed algorithm with state-of-the-art anomaly detection algorithms. In Section 5, we state the conclusions of this work and sketch future research directions.

\section{Background and Related Work}

\subsection{Anomaly Detection}

The main mechanism of anomaly detection algorithms is the construction of a model that determines a degree of ``normality" for the data points, and then detects anomalies as points that deviate from this model. The importance of anomaly detection lies in the fact that recognizing and understanding anomalous behavior allows one to make decisions and predictions. For example, anomalous traffic patterns on a network could mean that sensitive data is being transmitted across it, or that there is an intruder attempting to attack a system \cite{Bouyeddou2021}. Anomalous user behavior in payment transactions could signify credit card fraud and give some clues as to how to avoid it in the future \cite{Santosh2020}. Anomaly detection should be distinguished from outlier detection, the purpose of which is data cleaning: to locate outliers (mainly normal data affected by noise) and subsequently remove them from the data set.

Let $r$ be the ratio of anomaly samples with respect to all samples. When $r$ is high, the most common approach to anomaly detection is to use a supervised classification method. However, when $r$ is low (which is usually the case), the best approach is to use semi-supervised or unsupervised AD algorithms. According to \cite{Prasad2009}, classical AD algorithms can be classified into six main types: classification-based approaches (such as the well-known one-class SVM), probabilistic models (such as histograms and density estimation methods like KDE \cite{Kalair2021}), clustering models, information-based methods, spectral analysis (where dimensionality reduction methods such as PCA are found), and distance-based methods (including isolation forest and nearest neighbors).

In recent years, these classical models have been progressively replaced by deep learning-based algorithms, showing high performance in anomaly detection tasks \cite{Zhang2021}. These algorithms can efficiently model the inherent features of complex data through their distributed and hierarchical architectures, i.e., they can perform implicit feature engineering, especially when a large amount of data is available \cite{Rippel2021}. For this reason, they are commonly used for anomaly detection in scenarios involving very large data sets with high-dimensional instances, such as speech recognition, image/video analysis, or natural language processing \cite{pang2021deep}.

\subsection{Anomaly Detection Baseline Methods}
\label{subsec:baseline_methods}

In this paper, we select 11 state-of-the-art anomaly detection methods that represent a complete sample of the most common types of anomaly detection methods. All algorithms consider the proportion of anomalies in the data as a necessary parameter for finding threshold values; other parameters for specific algorithms are mentioned in later sections. 

We selected five well-known methods that are based on classic mathematical approaches to anomaly detection. These methods include One Class SVM \cite{scholkopf2001estimating}, a  kernel-based algorithm that encloses normal data in a boundary that leaves anomalies outside of it; Covariance Estimator, that finds the smallest ellipsoid that wraps normal data; Isolation Forest \cite{liu2008isolation}, that tries to separate points using decision trees, and those who are easier to isolate are the outliers; Local Outlier Factor (LOF) \cite{breunig2000lof}, based on a distance measure from each point to a set of its neighbors; and K-nearest neighbor (KNN) \cite{Ramaswamy2000}, that makes an scoring based on the distance from the $k$-th nearest neighbor.

We also decided to include methods developed in the last decade that do not use neural networks into their architectures, but instead take other approaches. These type of methods include SOS \cite{Janssens2012}, that builds an affinity matrix between all points that acts as a similarity measure; COPOD \cite{Li2020}, that builds univariant probability functions fro each dimension and then joins them in a unified multivariant function that models the true distribution of the data; and LODA \cite{Pevny2016}, that combines classifiers in low dimensions and uses histograms to detect anomalies.

Finally, we complete the overview of baseline methods with some of the proposals that are based on the use of neural networks as their central elements. These three models include VAE-Bayes \cite{Kingma2014}, built around a variational autoencoder that builds a latent space from data, where the probability distribution of the data can be retrieved by using Bayesian assumptions; DSVDD \cite{Ruff2018}, in where a neural network is used to transform data into a latent space where normal data can be encompasssed into a hypersphere whose features are neurally optimized; and LAKE \cite{Lv2020}, that includes a variational autoencoder to reduce dimensionality in a way that preserves data distribution, and then performs KDE in this space and separates anomalies using a threshold.

\section{Anomaly Detection through Density Matrices and Fourier Features (AD-DMKDE)}\label{section:anomaly_detection_method}

Gonz\'{a}lez et al. \cite{gallegoQNDE} proposed a new algorithm based on the joining of density matrices and random Fourier features for neural density estimation, called ``\textit{Density matrix for Kernel Density Estimation}" (DMKDE), that has its core in kernel density estimation. The DMKDE algorithm consists of a random Fourier feature mapping, a density matrix calculation, and a quantum density estimation phases. This method, whose original purpose was only density estimation, is used as the foundation for the novel method presented here.

\begin{figure*}
    \includegraphics[width=1.05\textwidth]{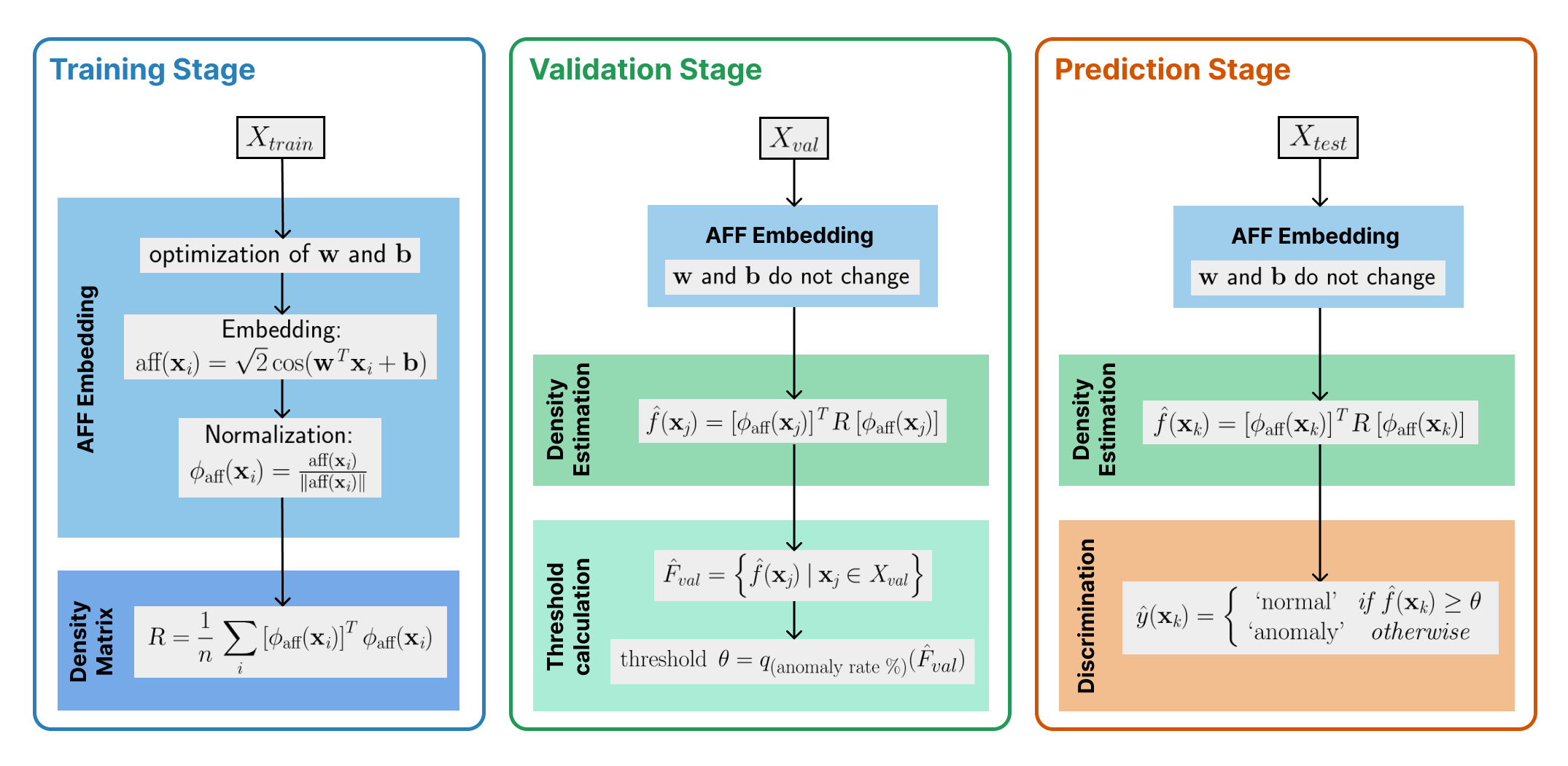}
    \caption{AD-DMKDE algorithm. Each section corresponds to a different stage of the method. The parameters of the mapping/embedding and the density matrix do not change after they are calculated. }
    
    \label{fig:AD-DMKDE_algorithm}
\end{figure*}

AD-DMKDE enhances the random Fourier feature mapping from DMKDE through the use of optimization, that allows to find better values for the parameters of the mapping function. This process, called Adaptive Fourier features, can be seen in more detail in subsection \ref{subsec:rff_and_aff}. After that, the density matrix is built from the mappings of the training samples (see subsection \ref{subsec:density_matrix}), and the density of new samples is calculated through a process known as Quantum Density Estimation, inspired in quantum measurement theory from quantum mechanics (see subsection \ref{subsec:qde}). Then, as final step, a threshold is obtained through a cross-validation process over a validation data set. This threshold is calculated using the rate of the anomalies in the validation set and is explained in the Subsection \ref{subsec:threshold_calculation}. The test phase of the algorithm comprises applying the already-built Fourier feature mapping over the test set, Quantum Density Estimation and threshold comparison to make the decision whether each test point is anomalous or not. Figure \ref{fig:AD-DMKDE_algorithm} shows a summary of the method.

\subsection{Random Fourier features and Adaptive Fourier features}
\label{subsec:rff_and_aff}

In the training phase, each data point in the training data set is implicitly transformed into a higher dimension. The Random Fourier feature mapping, as first stated in \cite{rahimi2007rff}, applies the following function $\text{rff}(\bf{x}_\textit{i}) = \sqrt{2} \cos(\bf{w}^\textit{T} \bf{x}_\textit{i} + \bm{b}) $ where $\bf{w}$ is a vector of $iid$ from a $\mathcal{N}(\bm{0},\bm{I}_D)$ and $\bf{b}$ is a $iid$ sample from a $\text{Uniform}(0,2\pi)$. According to Bochner's theorem, the kernel $k$ can be approximated as an expectation, $k(\bm{x},\bm{y}) \simeq \mathbb{E}_{\bm{w}}[\langle \phi_{\text{rff}}(\bm{y}), \phi_{\text{rff}}(\bm{x}) \rangle ]$. So, since we expect to approximate a Gaussian kernel to perform KDE, the parameters of the kernel, in this case the $\sigma$ (standard deviation) parameter, is one of the major influences over the selection of $\bf{w}$ and $\bf{b}$.

This mapping is improved through AFF (Adaptive Fourier features), a mechanism that, before of the RFF mapping, performs an optimization of the values of $\bf{w}$ and $\bf{b}$ by building a neural network that adjust their values through a learning process. After the application of the RFF formula with the optimized values of $\bf{w}$ and $\bf{b}$, the obtained results are normalized to obtain the final AFF embedding $  \phi_{\text{aff}}(\bf{x}_\textit{i})$ that will be the representation of each $\bf{x}_\textit{i}$.

\subsection{Density matrix calculation}
\label{subsec:density_matrix}

Once the AFF embedding has been applied to all training samples, the transformed training data correspond to the set $\{ \phi_{\text{aff}}(\bf{x}_\textit{1}) , \phi_{\text{aff}}(\bf{x}_\textit{2}) , ..., \phi_{\text{aff}}(\bf{x}_\textit{n}) \}$, 
being $n$ the number of train samples. For each sample, the algorithm computes its corresponding pure-state density matrix $\textsl{R}_i$, given by the outer product of each $\phi_{\text{aff}}(\bf{x}_\textit{i})$:

$$\textsl{R}_i = \left [ \phi_{\text{aff}}(\bf{x}_\textit{i}) \right ]^T \phi_{\text{aff}}(\bf{x}_\textit{i}) $$

The mixed-state density matrix that comprehends all the training samples is calculated as 
$\textsl{R} = 1/n \sum_{i=1}^{n} \textsl{R}_i$. Although $\textsl{R}$ contains all the information of the embedded train samples, its size does not depend on $n$; instead, it only depends on the size of the embedding. This is a major improvement over other methods, making it possible to compactly represent (and store) very large data sets using only a single matrix whose size is much smaller in general than the training set. Other advantage of the use of the density matrix representation is the fact that the calculation of $\textsl{R}$ only has to occur once in the entire training phase.

\subsection{Quantum Density Estimation}
\label{subsec:qde}

With the calculation of $\textsl{R}$, we can estimate the density for new samples, in particular those of the validation and test sets. Given a validation sample $\bf{x_\textit{j}}$, we need to transform it by using the exact same mapping and normalization we previously defined in order to obtain $\phi_{\text{aff}}(\bf{x_\textit{j}})$, and then calculate 

$$\hat{f}(\bf{x}_\textit{j}) = [ \phi_{\text{aff}}(\bf{x}_\textit{j}) ]^\textit{T} \, \textsl{R} \: [ \phi_{\text{aff}}(\bf{x}_\textit{j}) ]$$

as an estimate of the density of $\bf{x_j}$. This process is repeated for every validation sample, in order to build the validation density set 
$\hat{F}_{val} = \left\{ \hat{f}(\bf{x}_\textit{j}) \:|\: \bf{x}_\textit{j} \in \mathit{X_{val}} \right\}$.

\subsection{Threshold calculation and prediction phase}
\label{subsec:threshold_calculation}

 To determine which of the samples are anomalies or not, we need to find a threshold $\theta$ that can be used as a sort of boundary between these two types of data points. Therefore, we use the anomaly rate of the data set (either an a priori known value or the proportion of anomalies we expect to find). We take the estimates from the validation set and find the percentile that corresponds to the anomaly rate we defined above to calculate the threshold $\theta := q_{(\text{anomaly rate \%})} (\hat{F}_{\mathit{val}})$.

As a final step, we estimate the density of the test samples (by mapping them and applying density estimation with $\textsl{R}$), and then use the threshold $\theta$ to discriminate them. So, the classification of a test sample $\bf{x_k}$ is given by: 

$$\hat{y}(\bf{x}_\textit{k}) = \left\{\begin{matrix} \text{ `normal'} & \textit{if } \hat{f}(\bf{x}_\textit{k}) \geq \theta\\  \text{ `anomaly'} & \textit{otherwise}\end{matrix}\right.$$

\vspace{5pt}
\section{Experimental Evaluation}
\label{sec:experimental_evaluation}

\subsection{Experimental Setup}

For our experiments, we compared our proposed method with all the baseline algorithms listed above (see section \ref{subsec:baseline_methods}). To run the first four methods, we used the Python implementation provided by the scikit-learn library. KNN,  the shallow methods, VAE and DVSDD were executed through the implementation provided by the PyOD Python library \cite{Zhao2019}. For the LAKE algorithm \cite{Lv2020}, the implementation we used comes from the Github repository of its authors\footnote{https://github.com/1246170471/LAKE}, although we had to correct some issues in the original code, particularly the way to split the test data set to include both normal and anomalous samples.

To handle the inherent randomness that AD-DMKDE can present in some stages (particularly in AFF neural network training), we decided to select a unique, invariant value for every random seed that affect the behavior of the method. All the experiments were carried out on a machine with a 2.1GHz Intel Xeon 64-Core processor with 128GB RAM and two NVIDIA RTX A5000 graphic processing units, that run Ubuntu 20.04.2 operating system.

\subsubsection{Data sets.}

We used twenty public data sets to evaluate the performance of our proposed algorithm in performing anomaly detection. We chose data sets with a wide variety of characteristics, such as their size or anomaly rate, to test the robustness of our algorithm in multiple scenarios. The main characteristics of all the data sets can be seen in Table \ref{data set_features}. The source for all the data sets was the ODDS Library of Stony Brook University \cite{Rayana2016}.

\begin{table}[t]
\caption{Main features of the data sets.}
\label{data set_features}

\scriptsize
\centering
\begin{tabular}[c]{l|c|c|c}
\hline
Data Set & Instances & Dimensions & Outlier Rate \\ 
\hline

Arrhythmia & 452 & 274 & 0,146 \\
Cardio & 2060 & 22 & 0,2 \\
SpamBase & 3485 & 58 & 0,2 \\
Thyroid & 3772 & 36 & 0,0247 \\
KDDCUP & 10000 & 118 & 0,1934 \\
Glass & 214 & 9 & 0,042 \\
Lympho & 148 & 18 & 0,04 \\
Ionosphere & 351 & 33 & 0,359 \\
Letter & 1600 & 32 & 0,0625 \\
MNIST & 7603 & 100 & 0,092 \\
Musk & 3062 & 166 & 0,0317 \\
OptDigits & 5216 & 64 & 0,0288 \\
PenDigits & 6870 & 16 & 0,0227 \\
Pima & 768 & 8 & 0,349 \\
Satellite & 6435 & 36 & 0,3164 \\
SatImage & 5803 & 36 & 0,0122 \\
Shuttle & 10000 & 9 & 0,0715 \\
Vertebral & 240 & 6 & 0,125 \\
Vowels & 1456 & 12 & 0,03434 \\
WBC & 378 & 30 & 0,0556 \\  
\hline
\end{tabular}

\vskip -0.1in
\end{table}

\subsubsection{Experiment Configuration.}

For each data set, the following configuration was chosen: the data set was split in a stratified way (keeping the same proportion of outliers in each subset) by randomly taking 30\% of the samples as the test set, and from the remaining samples, again randomly taking 30\% for validation and the remaining 70\% as the training set. This splitting was performed only once per data set, so that all algorithms worked with exactly the same data partitions.

When implementing the Scikit-Learn and PyOD algorithms, the default configurations defined in these libraries were used, modifying only the parameters referred to the data set features mentioned in the previous section. The only exception to this rule was the VAE algorithm, whose architecture, based on autoencoders, needed to be slightly modified when using low-dimensional data sets, to ensure that the encoded representation was always of lower dimensionality than the original sample. Also, as mentioned above, LAKE algorithm required some corrections to the original code published by its authors; however, the basic internal structure, the internal logic of the algorithm and the functions on which it is based remained the same as in its original configuration.

Our AD-DMKDE algorithm was implemented from the original code of DMKDE, presented in \cite{gonzalez2022learning}, with some modifications. Since the main goal of DMKDE is density estimation, the separation of data as normal or outlier was performed by adding a function that calculates the classification threshold using the rate of anomalies in each data set, a known value in all cases (see table \ref{data set_features}).

\subsubsection{Evaluation Metrics.}

The main metric we chose to determine the performance of the algorithms was the F1 score (with weighted average), a well-known and very common metric used when testing classification algorithms. However, the calculation of other common metrics, such as accuracy and the exclusive F1-Score to the anomalous class, was also implemented. For each algorithm, a set of parameters of interest was selected in order to perform a comprehensive search for the combinations of values that gave the best result for each data set. This search was performed by training only with the training set and reporting the results of the validation set. The selection of the best combination of parameters was performed by choosing the model with the best average F1 score. Once the best parameters were found, these same values were used to run each algorithm once again, this time combining the training and validation sets for the training phase and reporting the results on the test set. These final results were the values we used to compare the overall performance of all algorithms.

\subsection{Results and discussion}

\begin{table*}[t]
\caption{F1 Score for all classifiers over all data sets. The first and second best values are marked in bold and underlined, respectively.}
\label{f1_score}

\scriptsize
\centering
\begin{tabular}[c]{l||c|c|c|c|c|c|c|c|c|c|c||c}
\hline
Data Set    & OCSVM & iForest & Cov. & LOF & KNN & SOS & COPOD & LODA   & VAE-B & DSVDD & LAKE   & AD-DMKDE \\
\hline

Arrhythmia & 0.813 & 0.821 & 0.818 & 0.804 & 0.861 & 0.773 & 0.844 & 0.798 & 0.856 & 0.864 & \underline{0.909} & \bf{0.911} \\
Cardio     & \underline{0.804} & 0.752 & 0.756 & 0.702 & 0.753 & 0.739 & 0.750 & 0.717 & 0.783 & 0.735 & 0.772 & \bf{0.831} \\
Glass      & 0.916 & 0.931 & 0.931 & 0.925 & 0.900 & 0.907 & 0.916 & 0.848 & 0.900 & 0.907 & \bf{1.000} & \underline{0.974} \\
Ionosphere & 0.765 & 0.710 & 0.876 & 0.830 & 0.817 & 0.784 & 0.736 & 0.510 & 0.714 & 0.674 & \bf{0.993} & \underline{0.959} \\
KDDCUP     & 0.744 & 0.838 & 0.975 & 0.789 & 0.932 & 0.716 & 0.785 & 0.770 & 0.803 & 0.755 & \bf{1.000} & \underline{0.984} \\
Letter     & 0.893 & 0.897 & 0.909 & \bf{0.930} & 0.910 & 0.911 & 0.895 & 0.899 & 0.897 & 0.897 & 0.838 & \underline{0.927} \\
Lympho     & 0.934 & \bf{1.000} & 0.934 & \bf{1.000} & \bf{1.000}  & 0.934 & 0.962 & 0.923 & \bf{1.000} & \bf{1.000}  & \bf{1.000} & \bf{1.000} \\
MNIST      & 0.881 & 0.881 & 0.841 & 0.886 & 0.909 & 0.864 & 0.868 & 0.866 & 0.895 & 0.880 & \bf{0.959} & \underline{0.911} \\
Musk       & 0.958 & 0.931 & \underline{0.997} & 0.958 & 0.991 & 0.951 & 0.964 & 0.954 & 0.984 & 0.992 & 0.639 & \bf{1.000} \\
OptDigits  & 0.952 & 0.952 & 0.952 & 0.949 & 0.952 & 0.953 & 0.949 & \underline{0.955} & 0.951 & 0.952 & 0.819 & \bf{0.981} \\
PenDigits  & 0.963 & \underline{0.967} & 0.966 & 0.961 & 0.965 & 0.963 & 0.966 & 0.966 & 0.966 & 0.963 & \bf{0.994} & \bf{0.994} \\
Pima       & 0.592 & 0.624 & 0.559 & 0.615 & 0.644 & 0.636 & 0.615 & 0.597 & 0.632 & 0.679 & \underline{0.740} & \bf{0.758} \\
Satellite  & 0.681 & 0.757 & 0.813 & 0.634 & 0.716 & 0.597 & 0.732 & 0.709 & 0.761 & 0.761 & \underline{0.841} & \bf{0.845} \\
SatImage   & 0.984 & \underline{0.998} & 0.991 & 0.979 & 0.998 & 0.980 & 0.994 & 0.997 & 0.996 & 0.996 & 0.946 & \bf{1.000} \\
Shuttle    & 0.933 & \underline{0.992} & 0.968 & 0.900 & 0.973 & 0.891 & 0.990 & 0.973 & 0.981 & 0.978 & 0.990 & \bf{0.998} \\
SpamBase   & 0.702 & 0.794 & 0.714 & 0.702 & 0.719 & 0.719 & 0.799 & 0.699 & 0.741 & 0.738 & \bf{0.885} & \underline{0.816} \\
Thyroid    & 0.953 & 0.958 & \bf{0.986} & 0.949 & 0.953 & 0.949  & 0.953 & 0.958 & 0.958 & 0.961 & 0.803 & \underline{0.967} \\
Vertebral  & 0.750 & 0.778 & 0.817 & 0.796 & 0.810 & 0.817 & 0.817 & 0.817 & 0.817 & \underline{0.819} & 0.807 & \bf{0.904} \\
Vowels     & 0.950 & 0.942 & 0.941 & 0.951 & 0.969 & 0.954 & 0.943 & 0.929 & 0.952 & 0.943 & \bf{1.000} & \underline{0.979} \\
WBC        & 0.942 & 0.941 & 0.949 & 0.957 & 0.942 & 0.913 & \bf{0.970} & 0.947 & 0.957 & 0.957 & 0.897 & \underline{0.961} \\
\hline
\end{tabular}

\end{table*}

The results for all data sets and all algorithms (baseline methods and our proposed method) are shown in Table \ref{f1_score}. The best value for each data set is highlighted in bold, and the second best value is underlined. In the F1 Score table, a noticeable difference appears between most of the baseline methods and our proposed method, with LAKE being a notable exception; however, AD-DMKDE shows the best performance in the majority of the data sets, followed by LAKE and LOF methods.  

When we consider the influence of the characteristics of the data set on the performance of our method, we see an interesting pattern. The outlier rate of the data sets does not seem to have any influence over the performance of AD-DMKDE, since the method performs well when there are a low number of anomalies (like in Musk or Shuttle), and also when there are many of them (like in Cardio or Pima). The number of dimensions have a weak influence, since the method has the best performance more frequently in data sets with less than twenty dimensions (such as in Vertebral, Pima or Pendigits), but also in half of the data sets with more than 100 dimensions (Arrhythmia and Musk). Finally, when we look at the number of samples, we see that AD-DMKDE stands out more for data sets with few instances (less than 1000) , and also for many instances (more than 5000), but its performance is not as good in the intermediate data sets (the ones that have between 1000 and 5000 instances). In summary, AD-DMKDE is a method that can perform over the average state-of-the-art anomaly detection methods, standing out in many different scenarios and being capable of affording data sets with both low and high values in its number of dimensions, number of samples and outlier rates.

\section{Conclusion}
\label{sec:conclusions}

In this paper, we presented a novel approach for anomaly detection using density matrices from quantum mechanics and random Fourier features. The new method AD-DMKDE was systematically compared against eleven different anomaly detection algorithms using F1 Score as main metric. AD-DMKDE show better than state-of-the-art performance over twenty anomaly detection data sets, being notably superior than classic algorithms and comparable to deep learning-based methods. The performance of the method does not seem to be affected by the anomaly rate of the data sets or the size of the data sets, but it seems to perform better for data sets with low dimensionality of the samples. AD-DMKDE do not have huge memory requirements due to that the method constructs a single density matrix in all training phase whose size is defined by the embedding. So it is possible to resume very large data sets in relatively small matrices, which makes AD-DMKDE a scalable solution in those cases. Also, the method allows easy interpretability of the results, because each data point labeled as ``anomaly" can be understood as a sample lying in low density regions with respect to normal data. As future work, we will continue to further develop the main concepts of AD-DMKDE, building new algorithms based on the combination of Fourier features and density matrices with deeper neural networks, such as autoencoders and variational autoencoders, whose reduction power can be coupled to the new method in order to better handle data sets with higher dimensionalities.

\bibliography{anomaly_detection_2022}
\bibliographystyle{splncs04}

\end{document}